\documentclass{article}

\usepackage[preprint]{corl_2026} 

\usepackage{graphicx, booktabs, amsmath, multirow, xcolor, mwe, bm, amssymb, wrapfig, tabularx, caption, etoc}

\title{VLK: Learning Humanoid Loco-Manipulation from Synthetic Interactions in Reconstructed Scenes}

\definecolor{citecolor}{HTML}{0071bc}
\author{
\textbf{Yen-Jen Wang}$^{*,1,2}$,
\textbf{Jiaman Li}$^{*,\ddagger,1}$,
\textbf{Sirui Chen}$^{\S,1,3}$,
\textbf{Takara E. Truong}$^{\S,1,3}$,
\textbf{Pei Xu}$^{\S,1}$
\\
\textbf{Pieter Abbeel}$^{\dagger,1,2}$,
\textbf{Rocky Duan}$^{\dagger,1}$,
\textbf{Koushil Sreenath}$^{\dagger,1,2}$,
\textbf{Angjoo Kanazawa}$^{\dagger,1,2}$
\\
\textbf{Carmelo Sferrazza}$^{\dagger,1}$,
\textbf{Guanya Shi}$^{\dagger,1,4}$,
\textbf{C. Karen Liu}$^{\dagger,1,3}$
\\[0.5em]
$^{1}$Amazon FAR \quad
$^{2}$UC Berkeley \quad
$^{3}$Stanford University \quad
$^{4}$Carnegie Mellon University
}

\definecolor{gold}{RGB}{255,165,0}
\definecolor{maroon}{RGB}{128,0,0}

\newcommand{\website}{https://vision-language-kinematics.github.io}

\begin{document}
\maketitle

\begingroup
\renewcommand\thefootnote{}
\footnotetext{
$^{*}$Co-first authors.
$^{\ddagger}$Project lead.
$^{\S}$Equal contribution.
$^{\dagger}$Amazon FAR Team Co-Lead.
}
\endgroup

\vspace{-8mm}
\begin{center}
\includegraphics[width=\textwidth]{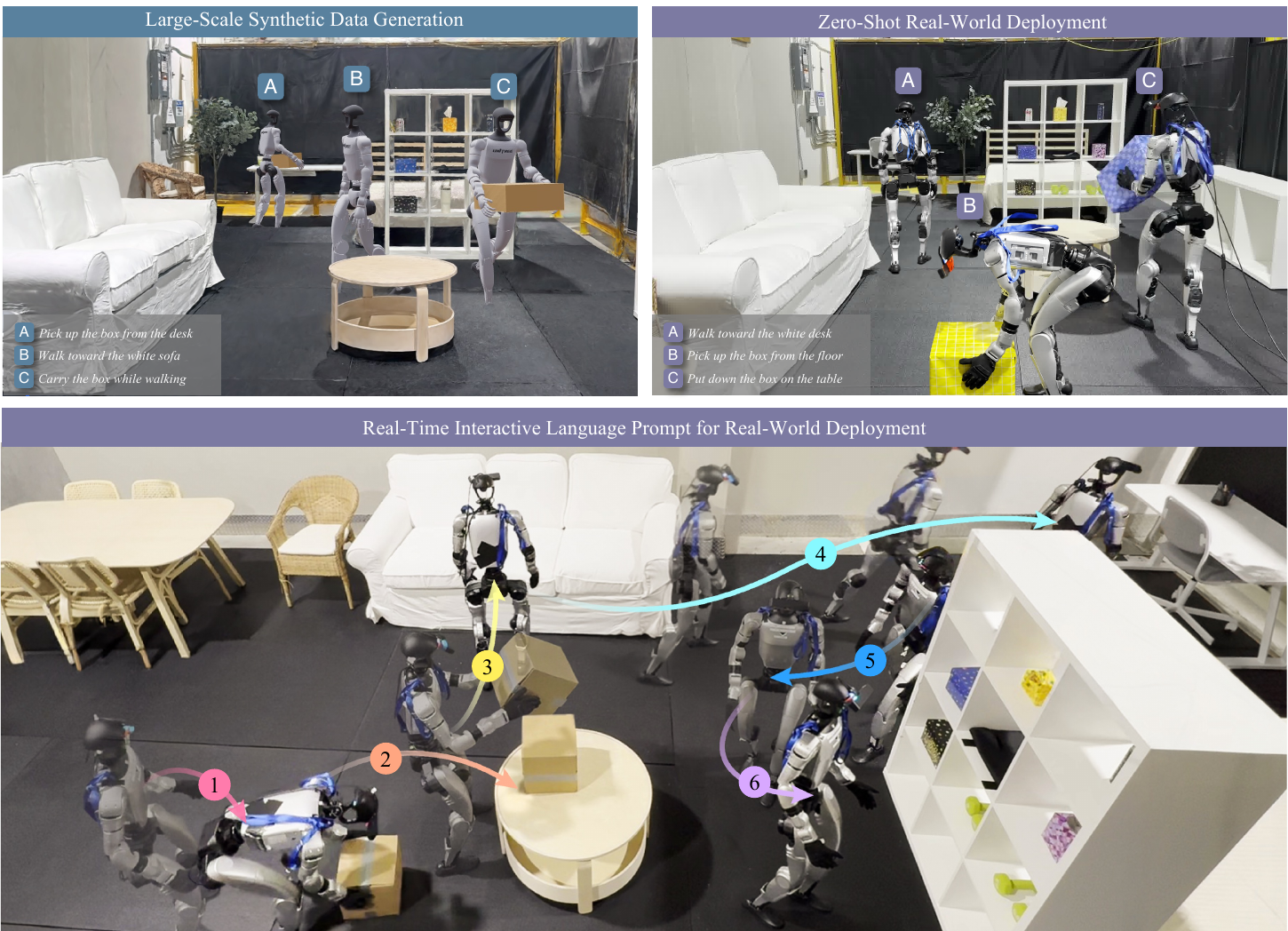}

\captionof{figure}{
\small
\textbf{Synthetic interactions in reconstructed scenes enable real-world perception-based loco-manipulation.}
Synthetic data generation produces paired egocentric observations, task instructions, and G1 kinematic trajectories for VLK training.
\textbf{Top:} A--C show atomic prompt-conditioned skills generated in \textbf{reconstructed scenes (left)} and executed on the physical G1 in the real world (right).
\textbf{Bottom}: long-horizon real-world behavior is produced by chaining atomic prompts; markers 1--6 indicate successive stages for picking, placing, and navigation. Project Website: \href{\website}{\textcolor{citecolor}{\website}}.
}
\label{fig:teaser}
\vspace{-3mm}
\end{center}

\begin{abstract}
Perception-based humanoid loco-manipulation requires connecting egocentric observations and task instructions to whole-body motion. Learning this mapping requires synchronized egocentric images, language commands, and robot-compatible kinematic trajectories, yet no existing data source provides this complete tuple at scale. We address this bottleneck by generating vision-language-kinematics (VLK) supervision synthetically in reconstructed scenes. Our pipeline leverages 3D Gaussian Splatting to reconstruct metric-scale indoor environments, synthesizes navigation and object-interaction trajectories using privileged scene information, and renders paired egocentric observations after the fact. We produce 48,000 paired trajectories with no human intervention and train a VLK policy that predicts short-horizon whole-body kinematic trajectories. A whole-body tracker converts these predictions into actions on the physical humanoid. We evaluate on the physical Unitree G1 performing navigation and single-object transport, demonstrating that synthesized interactions in reconstructed scenes provide effective supervision for sim-to-real perception-based humanoid loco-manipulation.
\end{abstract}

\etocdepthtag.toc{main}
\vspace{-10pt}
\section{Introduction}\vspace{-10pt}

Connecting egocentric observations to whole-body action is a fundamental challenge for humanoid robots operating in human-centered environments. A humanoid must perceive the world from its own viewpoint, identify task-relevant objects, navigate toward them, and physically interact with them through coordinated full-body motion. This perception-to-action loop, which humans execute effortlessly in everyday settings, requires a policy that maps high-dimensional visual input and language instructions to whole-body kinematic behavior.

Learning such a policy requires paired data: synchronized egocentric observations, task instructions, and robot-compatible whole-body trajectories. Prior work has explored several paths to obtain this supervision, each with significant limitations. Real-world teleoperation systems \cite{ze2025twist2,li2025clone} can produce high-quality paired demonstrations, but collecting full-body data remains expensive and difficult to scale across diverse scenes, objects, and interaction types. Human motion-capture datasets offer rich whole-body references that can be retargeted to humanoid morphologies, but they lack the corresponding robot egocentric observations. Egocentric video datasets provide abundant visual experience, but are not paired with robot-compatible kinematics or actions. No single existing source provides the complete tuple — vision, language, and whole-body kinematics — needed for scalable policy learning.

Can we overcome this data bottleneck by generating paired supervision \emph{synthetically}? Doing so requires two ingredients: \emph{synthetic world assets} that are visually realistic and diverse, and \emph{synthetic robot behaviors} that produce valid whole-body trajectories within those worlds. Each ingredient poses a distinct challenge. On the Real2Sim side, creating environments with sufficient realism, diversity, and geometric detail at scale has traditionally been prohibitive. On the Sim2Real side, policies trained entirely on synthetic renderings must transfer back to physical hardware under real-world visual and dynamic variation. Recent progress, however, suggests that these barriers are beginning to weaken. 3D Gaussian Splatting (3DGS) \cite{3dgs} can efficiently reconstruct indoor environments with photorealistic detail and metric scale, providing a partial Real2Sim solution for world assets. Whole-body motion tracking trained in simulation \cite{liao2025beyondmimic} can produce real-world robot policies that faithfully execute kinematic reference trajectories, providing a partial Sim2Real solution for behaviors. From this maturing but still fragmented technical landscape, we ask: can we connect these partial solutions into a real-to-sim-to-real pipeline for both scene assets and humanoid behaviors?

We show that the missing link is a data-generation system that synthesizes paired visual observations within 3DGS worlds and corresponding robot kinematic behaviors suitable for whole-body motion tracking. The key advantage of such a system is the decoupling it enables. We can prescribe the language commands, leverage privileged information in the simulated world (e.g., object poses, collision geometry, walkable regions) to ease the challenge of behavior generation, and render the corresponding egocentric observations as RGB images after the fact. By exploiting targeted annotation, oracle-conditioned motion synthesis, and hindsight rendering, we can scale up data production by orders of magnitude to generate synchronized language instructions, egocentric visual observations, and whole-body kinematic trajectories without any human demonstration effort.


Using this system, we synthesize 48,000 trajectories automatically within 600 GPU-hours in metric-scale indoor environments reconstructed by 3DGS. We then train a Vision-Language-Kinematics (VLK) policy that takes an egocentric image, a task instruction, and the current robot state as input, and predicts a short-horizon whole-body kinematic trajectory together with binary wrist-object contact labels. We treat these contact labels as auxiliary components of the kinematic state: they indicate whether each wrist should be in contact with the manipulated object and are readily available from our synthetic data. During deployment, a contact-aware whole-body tracker~\cite{chen2026scenebot} uses the predicted kinematics and contact labels to produce robot actions on the physical humanoid.


We evaluate the system on two fundamental task categories, navigation and object transport. Navigation requires the robot to move itself to a commanded location; object transport requires it to move an object to a commanded location. Together, they exercise the full perception-to-action loop: visual grounding, locomotion, picking up, carrying, and placing objects. Experiments in closed-loop simulation and on the physical G1 demonstrate that our synthesized interactions in reconstructed scenes provide effective supervision for sim-to-real perception-based humanoid loco-manipulation.

\vspace{-10pt}

\section{Related Work} \vspace{-10pt}

\paragraph{Vision-language and perception-based policies for humanoids.}
Vision-language-action (VLA) policies map visual observations and language instructions to robot actions and have become a promising paradigm for robot policy learning~\cite{brohan2023rt2,kim2024openvla,black2024pi0}. Recent work extends this direction to humanoid robots, where policies must connect perception, language, and whole-body control~\cite{bjorck2025groot,xue2025leverb,jiang2025wholebodyvla,ding2025humanoid,wei2026psi0,chen2025head,shao2025langwbc,li2026fromw1,kalaria2025dreamcontrol}. 
WholeBodyVLA~\cite{jiang2025wholebodyvla} and $\Psi_0$~\cite{wei2026psi0} train humanoid loco-manipulation policies using egocentric video data together with real humanoid teleoperation data. Our work is complementary: instead of relying on existing robot demonstrations or human videos as the primary supervision source, we generate paired vision-language-kinematics supervision in reconstructed scenes, enabling a VLK policy to predict explicit future G1 whole-body kinematic trajectories for tracker-based execution.
\vspace{-3mm}
\paragraph{Synthetic data and reconstructed scenes for robot learning.}
Synthetic data can reduce real-world data-collection cost and expose policies to controlled variation in appearance, geometry, and task configuration. Recent work uses digital twins and reconstructed scene representations, such as 3D Gaussian Splatting, to build realistic synthetic environments for robot manipulation, locomotion, demonstration synthesis, and evaluation~\cite{robotwin,splatsim,robogsim,rlgsbridge,gaussgym}. In this work, we also use 3DGS-reconstructed scenes to generate synthetic robot-learning data. Our focus, however, is humanoid loco-manipulation, where data generation must provide both realistic egocentric observations and valid whole-body interaction behaviors. We therefore combine reconstructed scenes with humanoid motion synthesis to generate paired vision-language-kinematics supervision.

\vspace{-3mm}
\paragraph{Humanoid motion synthesis and whole-body tracking.}
Human motion, retargeting pipelines, and whole-body tracking controllers provide a practical route for generating and executing humanoid kinematic references. Retargeting methods map human motions to humanoid embodiments~\cite{yang2025omniretarget,araujo2025gmr}, and whole-body tracking methods train controllers to execute such references on hardware~\cite{he2024omnih2o,chen2025gmt,truong2025beyondmimic,zhao2025resmimic,luo2025sonic}. Recent work further generates text-conditioned humanoid motion references for whole-body control~\cite{dreamcontrol,zhang2026wholebodylocomotion,li2026fromw1,shao2025langwbc,rempe2026kimodo}. For humanoid loco-manipulation, however, synthetic behaviors must include object interaction. We therefore adapt human-object interaction synthesis methods~\cite{li2024chois,wu2025hihi} from SMPL-based human motion to a Unitree G1 kinematic representation, and use the synthesized G1-object trajectories as supervision paired with egocentric renderings and language instructions.

\vspace{-8pt}

\section{Method}
\begin{figure*}[t]
    \centering
    \includegraphics[width=\textwidth]{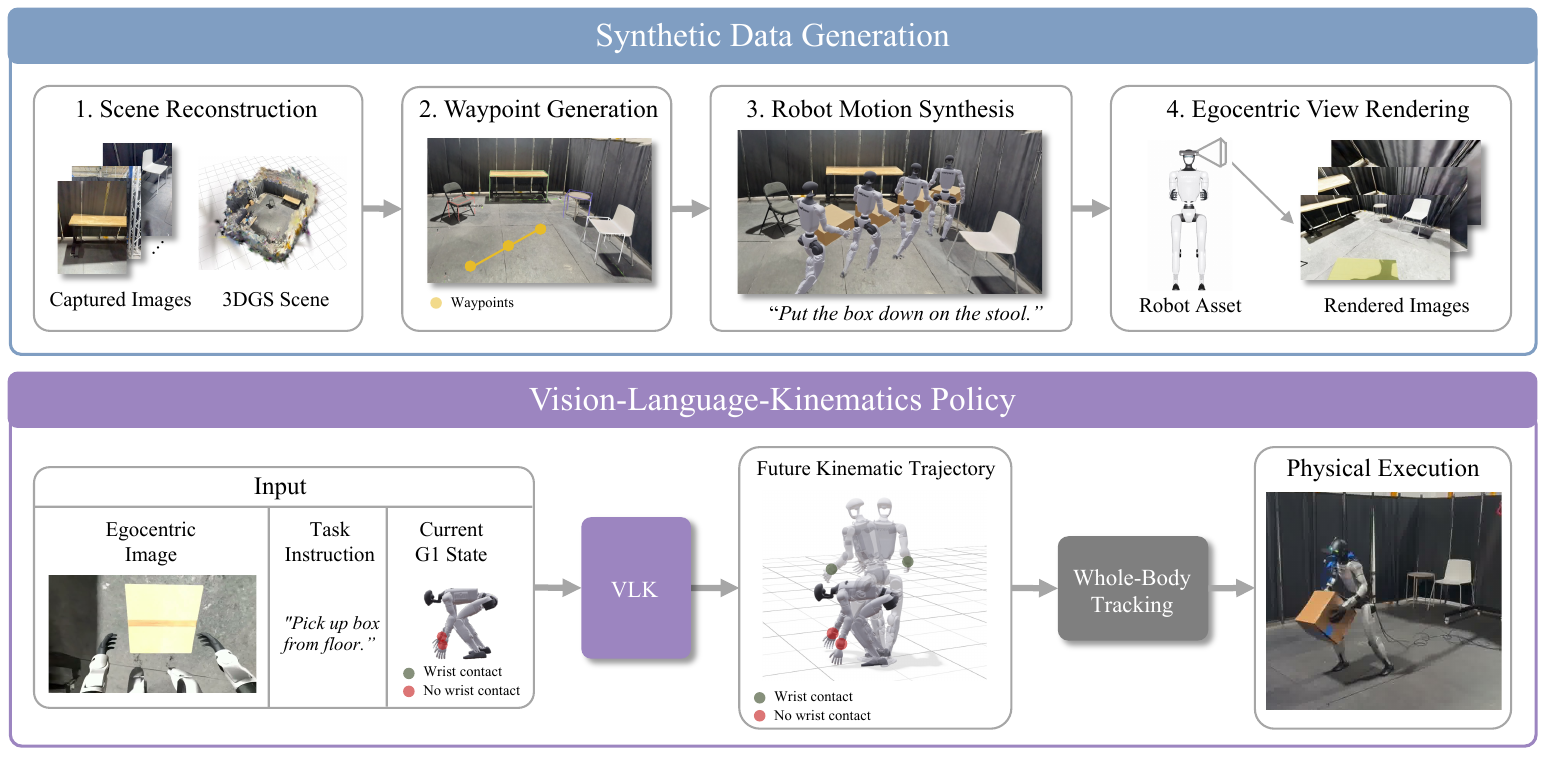}
\vspace{-6mm}
\caption{\textbf{Method overview.}
Top: we reconstruct 3D scenes, generate task waypoints, synthesize G1 motions, and render egocentric observations to produce paired vision-language-kinematics supervision.
Bottom: the paired data train a VLK policy to predict short-horizon G1 kinematic trajectories from egocentric observations, task instructions, and the current G1 state; at deployment, a whole-body tracker converts the predicted trajectories into robot actions for physical execution.
}
    \label{fig:method_overview}
    \vspace{-6mm}
\end{figure*}

\vspace{-10pt}
Our goal is to enable a Unitree G1 humanoid to perform perception-based loco-manipulation, including object-directed navigation and box interaction, from egocentric observations and task instructions. We formulate the problem as kinematic prediction followed by whole-body tracking: given the current egocentric view, instruction, and robot kinematic state, a high-level policy predicts a short-horizon G1 whole-body kinematic trajectory, and a whole-body tracker converts the predicted trajectory into executable robot actions. This decomposition makes whole-body kinematics the learning target for the perception policy, and shifts the central data requirement to paired egocentric observations, task instructions, and G1 kinematics. We obtain this supervision synthetically in reconstructed scenes, avoiding the need to collect full-body teleoperation data for the target tasks. Figure~\ref{fig:method_overview} illustrates the overall pipeline. The following sections describe synthetic data generation (Sec~\ref{synthetic_data_generation}), VLK policy training (Sec~\ref{vlk_policy}), whole-body tracking (Sec~\ref{whole_body_tracking}), and deployment (Sec~\ref{deployment}).

\vspace{-5pt}

\subsection{Synthetic Data Generation in Reconstructed Scenes}
\label{synthetic_data_generation}
\vspace{-5pt}

We generate training data by synthesizing Unitree G1 motions in metric-scale reconstructed scenes and rendering the corresponding egocentric observations. The pipeline has three stages: scene reconstruction and annotation, G1 motion synthesis in the annotated scenes, and egocentric rendering. Together, these stages produce paired vision-language-kinematics data.
\vspace{-3mm}
\paragraph{Scene Reconstruction and Annotation.}
We reconstruct indoor environments from Polycam scans captured with an iPhone 14 Pro. The scans combine RGB imagery with LiDAR-based depth, enabling metric-scale scene reconstruction. For each scene, we optimize a 3DGS representation that preserves the visual appearance and spatial layout of the real environment for egocentric rendering.

For motion synthesis, we extract a scene point cloud and annotate task-relevant semantic and geometric information. Specifically, we annotate semantic 3D bounding boxes for objects and mark walkable regions (see Appendix~\ref{app:scene_annotation}). These annotations define the scene constraints used by the motion synthesis stage: for each task, we sample task-relevant objects, feasible initial robot poses, and sparse waypoints according to task-specific rules. We also use semantic object labels from the 3D bounding boxes to instantiate task instructions with simple templates, e.g., ``walk toward the [object]'' for navigation and task-specific templates such as ``pick up the box from the floor'' or ``put down the box on the [surface]'' for interaction. Additional details are provided in Appendix~\ref{app:task_generation}.


\vspace{-4mm}
\paragraph{Interaction Synthesis in Reconstructed Scenes.}
Using the scene-derived waypoints, we synthesize Unitree G1 trajectories for two task families: object-directed navigation and object interaction. The navigation model is trained on G1 motions from BONES-SEED~\cite{bones2025seed}, while the interaction model is trained from OMOMO~\cite{li2023omomo} sequences retargeted to G1 with OmniRetarget~\cite{yang2025omniretarget}.

We represent a G1 motion sequence of length $T$ as
$\mathbf{X} = [\mathbf{p}, \mathbf{R}, \sin(\mathbf{q}), \cos(\mathbf{q})]$,
where $\mathbf{p} \in \mathbb{R}^{T \times J \times 3}$ and
$\mathbf{R} \in \mathbb{R}^{T \times J \times 6}$ denote global joint positions and 6D joint rotations~\cite{zhou2019continuity} for $J$ joints, and
$\mathbf{q} \in \mathbb{R}^{T \times D}$ denotes the G1 joint angles. We encode joint angles using $(\sin(\mathbf{q}), \cos(\mathbf{q}))$ to avoid discontinuities from angle wrap-around. For object-interaction sequences, we additionally represent the object trajectory as
$\mathbf{X}^{\mathrm{obj}} = [\mathbf{o}, \mathbf{R}^{\mathrm{obj}}]$,
where $\mathbf{o} \in \mathbb{R}^{T \times 3}$ is the global object position and
$\mathbf{R}^{\mathrm{obj}} \in \mathbb{R}^{T \times 6}$ is the object rotation relative to the input object geometry.

We build on the conditional diffusion formulation of prior human-object interaction synthesis methods~\cite{li2024chois,wu2025hihi}, and adapt it to the Unitree G1 representation. The navigation model generates G1 trajectories for object-directed walking from task instructions, initial G1 states, and sparse scene-grounded waypoints. The interaction model generates coupled G1-object trajectories and additionally conditions on the initial object state, object geometry features~\cite{prokudin2019efficient}, desired relative wrist poses, and wrist contact labels. These conditions specify interaction constraints such as wrist placement, and contact timing. Additional details are provided in Appendix~\ref{app:interaction_synthesis} and~\ref{app:motion_postprocessing}.

\vspace{-4mm}
\paragraph{Egocentric View Rendering.}
We render egocentric observations by replaying the synthesized G1 trajectories in Isaac Sim. For each sequence, we load the reconstructed 3DGS scene and place the Unitree G1 model in the scene. The robot is equipped with a virtual ZED~2i camera mounted on the head, with extrinsics calibrated to match the physical camera setup on the real robot. This alignment ensures that the rendered observations closely match the deployment viewpoint. To improve robustness to visual differences between rendered and real observations, we apply domain randomization during rendering, including lighting variation and perturbations to camera extrinsics and focal length. Details are provided in Appendix~\ref{app:domain_randomization}.

Overall, the pipeline produces paired VLK sequences consisting of egocentric observations, task instructions, and G1 whole-body trajectories. Examples of the resulting data are shown in Figure~\ref{fig:dataset_examples_box}. For more examples, please see Appendix~\ref{app:dataset_examples}.

\begin{figure*}[t]
    \vspace{-20pt}
    \centering
    \includegraphics[width=\textwidth]{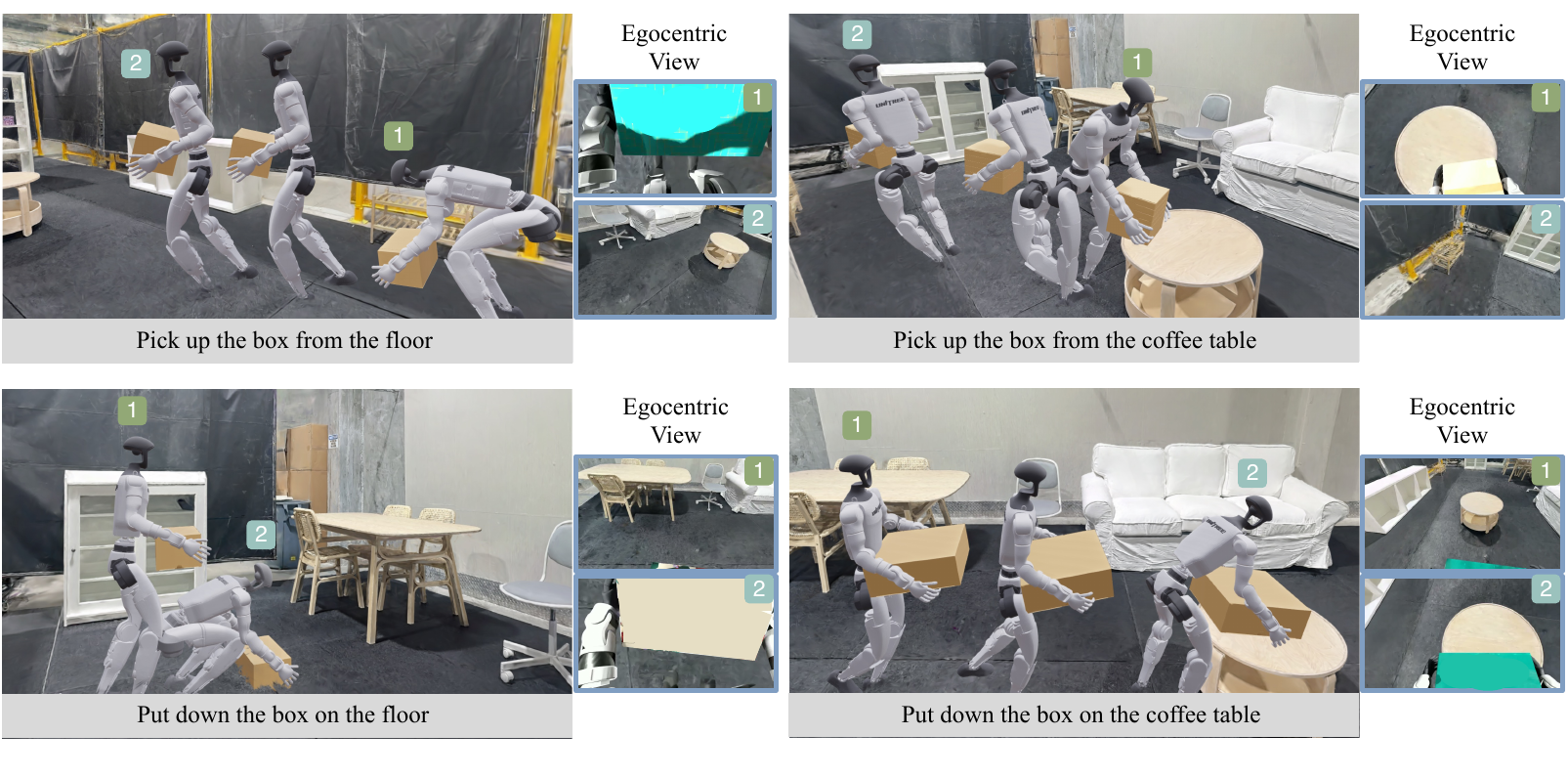}
    \vspace{-8mm}
  \caption{
\textbf{Examples of generated VLK supervision.}
Each sequence pairs an egocentric RGB observation and language instruction with the corresponding G1 whole-body kinematic trajectory.
}
    \label{fig:dataset_examples_box}
    \vspace{-5mm}
\end{figure*}

\vspace{-8pt}
\subsection{Vision-Language-Kinematics Policy}
\label{vlk_policy}
\vspace{-8pt}

The generated vision-language-kinematics sequences provide paired supervision for learning a policy that maps robot-view observations and task instruction to future whole-body motion.

\vspace{-3mm}
\paragraph{Kinematic State Representation.}
We represent each per-frame G1 kinematic state at frame $\tau$ as $\mathbf{x}_\tau=[\Delta x_\tau,\Delta y_\tau,\Delta\psi_\tau,h_\tau,\mathbf{R}^{\mathrm{root}}_\tau,\sin(\mathbf{q}_\tau),\cos(\mathbf{q}_\tau),\mathbf{c}_\tau]$, where $(\Delta x_\tau,\Delta y_\tau)$ is the heading-normalized root planar displacement, $\Delta\psi_\tau$ is the yaw change, $h_\tau$ is the root height, $\mathbf{R}^{\mathrm{root}}_\tau$ is the heading-normalized 6D root orientation, and $\mathbf{q}_\tau$ denotes the G1 joint angles. We denote the contact vector as $\mathbf{c}_\tau=[c_\tau^{L},c_\tau^{R}]$, where $c_\tau^{L},c_\tau^{R}\in\{0,1\}$ indicate the left and right wrist-object contact labels.

\vspace{-3mm}
\paragraph{VLK Policy Formulation.}
At timestep $t$, the VLK policy takes the current egocentric RGB observation $o_t$, task instruction $\ell$, and current G1 kinematic state $\mathbf{x}_t$ as input~\footnote{\vspace{-2mm}During deployment, the contact label in $\mathbf{x}_t$ is auto-regressively predicted rather than directly observed.}, and predicts an $H$-frame future trajectory $\hat{\mathbf{x}}_{t+1:t+H}=\pi_\theta(o_t,\ell,\mathbf{x}_t)$, where $H=30$ corresponds to one second at 30\,Hz. The instruction $\ell$ is fixed for each task execution, and the predicted trajectory serves as the high-level motion target for the downstream whole-body tracker.

\vspace{-3mm}
\paragraph{Training Objective.}
We initialize the VLK policy from a pretrained $\pi_{0.5}$~\cite{intelligence2025pi_}
and fine-tune the full model on our generated vision-language-kinematics dataset. We adapt the action space to our G1 kinematic representation. We train the policy with an $x_0$-prediction objective in a flow-matching formulation. Given a ground-truth future trajectory $\mathbf{x}_{t+1:t+H}$, Gaussian noise $\boldsymbol{\epsilon}$, and interpolation coefficient $\alpha$, we construct the noisy trajectory $\mathbf{x}^{\alpha}_{t+1:t+H}=\alpha\boldsymbol{\epsilon}+(1-\alpha)\mathbf{x}_{t+1:t+H}$. The policy predicts the clean trajectory as $\hat{\mathbf{x}}_{t+1:t+H}=\pi_\theta(\mathbf{x}^{\alpha}_{t+1:t+H},\alpha,o_t,\ell,\mathbf{x}_t)$ and is trained with the reconstruction loss $\mathcal{L}_{\mathrm{traj}}=\|\hat{\mathbf{x}}_{t+1:t+H}-\mathbf{x}_{t+1:t+H}\|_2^2$. We additionally apply auxiliary losses for foot-floor contact prediction, accumulated root-trajectory consistency, forward-kinematics accuracy, and foot-sliding regularization to improve motion quality. Details are provided in Appendix~\ref{app:vlk_training_details}.

\vspace{-5pt}
\subsection{Whole-Body Motion Tracking}
\label{whole_body_tracking}
\vspace{-5pt}

The VLK policy predicts high-level reference states $\hat{\mathbf{x}}_t$, which are executed on the physical humanoid by a contact-aware whole-body tracker based on SceneBot~\cite{chen2026scenebot}. The tracker is blind to egocentric observations and language instructions: it only receives converted reference targets and current robot proprioception, while VLK handles perception-conditioned replanning. At each control step, we convert the predicted VLK state $\hat{\mathbf{x}}_t$ into the tracker reference format, including lower-body joint targets, upper-body head and wrist target 6D poses, root target pose, and binary wrist-object contact labels. The tracker takes this converted reference $\bar{\mathbf{x}}_t$ and the current low-level robot state $\mathbf{s}_t$ as input, and outputs joint-level PD targets for all actuated joints:
$\mathbf{u}_t=\pi_{\mathrm{track}}(\bar{\mathbf{x}}_t,\mathbf{s}_t)$.
Similar to prior whole-body tracking controllers~\cite{chip,chen2026scenebot}, the tracker follows the reference motion while using the predicted contact labels to stabilize object transport. When wrist contact is active, the tracker activates contact-aware wrist behavior to maintain contact with the manipulated object.




\vspace{-5pt}

\subsection{Deployment}
\label{deployment}
\vspace{-5pt}

During real-world deployment, we implement a real-time VLA~\cite{ma2025running} system with Triton as the optimized inference backend. On an RTX~5090, the end-to-end VLK inference latency remains stable at $31\,\mathrm{ms}$. To reduce discontinuities between predicted chunks, we adapt the client-side real-time chunking idea from $\Psi_0$~\cite{wei2026psi0} to our kinematic trajectory setting. In our case, there will be $10$ time-frame overlap between two adjacent action chunks.
In addition, motion blur in egocentric observations can degrade VLK prediction quality during real-world operation, especially during fast bending motions. To improve robustness, we use sharpness-based frame selection during deployment. Details are provided in Appendix~\ref{app:motion_blur_deployment_details}.

\vspace{-5pt}

\section{Experiments}
\label{sec:experiment}
\vspace{-5pt}

In this section, we introduce our hardware setup (Sec.~\ref{subsec:experimental_setup}) and then answer four research questions:
(i) How efficiently can the proposed pipeline generate diverse paired VLK data? (Sec.~\ref{subsec:syn_data_gen})
(ii) Can a VLK policy trained on synthesized data transfer to the physical humanoid? (Sec.~\ref{subsec:evaluation_complete_system})
(iii) How does the amount of synthesized training data affect performance across different tasks? (Sec.~\ref{subsec:data_scale_ablation_study})
(iv) How does 
visual domain randomization help bridge the visual sim-to-real gap? (Sec.~\ref{subsec:domain_rand_rendering})

\vspace{-5pt}

\subsection{Experimental Setup}
\label{subsec:experimental_setup}
\vspace{-5pt}
We use two physical environments: a lab-style scene and an apartment-style scene. For each environment, we create multiple physical layouts by changing the positions and orientations of selected furniture items, then scan each layout using the Polycam app on an iPhone~14 Pro and reconstruct it with 3D Gaussian Splatting (3DGS)~\cite{3dgs,gaussgym}. This produces 4 lab and 4 apartment layouts, which are used for synthetic data generation and simulation evaluation. Real-world deployment is conducted in the same two physical environments, with furniture layouts manually varied across trials.

We deploy our system on a Unitree G1 
on a custom 3D-printed head bracket to obtain a wider field of view and use the left RGB image at 672$\times$376 resolution, resized to the VLK input resolution of 224$\times$224. During deployment, VLK model inference runs on an RTX~5090, while the whole-body tracking policy runs on an RTX~5000 Ada. A detailed system overview and profiling results are provided in Appendix~\ref{app:deployment_profiling}.
\vspace{-5pt}

\subsection{Synthetic Data Generation}
\label{subsec:syn_data_gen}
\vspace{-5pt}
\paragraph{How efficiently can the proposed pipeline generate diverse paired VLK data?}
Given a reconstructed scene with semantic 3D bounding boxes and walkable-region annotations, the pipeline samples task-relevant objects, feasible initial poses, waypoints, and task instructions; synthesizes G1-compatible trajectories; filters trajectories with scene penetrations; and renders paired egocentric observations. Each sequence contains synchronized egocentric observations, language instructions, and G1 whole-body kinematic trajectories.

Each environment contains 4 reconstructed layouts and 12 data-generation modes: six evaluated modes---Walk To, Turn Around, Pick (Floor), Put (Floor), Pick (Surface), and Put (Surface)---plus auxiliary modes for random walking, turning, carrying, and carry-turning. These auxiliary modes improve state coverage by exposing the policy to diverse walking and object-transport behaviors. We synthesize 1000 trajectories per layout and mode, yielding 48,000 trajectories per environment. On a single NVIDIA L40S GPU, synthesizing 1000 trajectories for one mode in one layout takes approximately 4 hours, while rendering the corresponding egocentric observations takes approximately 8.3 hours. These results show that, after scene reconstruction and annotation, the pipeline can automatically generate large volumes of paired VLK supervision, with generation parallelized across layouts and modes.
\vspace{-12pt}

\subsection{Evaluation of Complete System}
\label{subsec:evaluation_complete_system}
\vspace{-8pt}

\paragraph{Can a VLK policy trained on synthesized data transfer to the physical humanoid?}
We test the full closed-loop system in MuJoCo simulation (with IsaacSim rendering) and on the physical Unitree G1. In both settings, the VLK policy predicts short-horizon whole-body kinematic trajectories from egocentric observations, task instructions, and robot kinematic state, while the whole-body tracker converts the predicted kinematics into robot actions.

For real-world evaluation, we conduct trials in both the lab and apartment environments while varying the scene layout across trials. We evaluate the six task modes shown in Table~\ref{tab:eval_system}, with 20 trials per mode, resulting in 120 real-world trials in total. In addition, we also evaluate ``Pick (Floor)" without a contact label, and the success rates in both the Lab Scene and the Apartment Scene are 0 out of 5 trials in real-world evaluation. 
For simulation evaluation, we use Isaac Sim to render egocentric RGB observations from the reconstructed 3DGS scenes. At each replanning step, the VLK policy takes the rendered egocentric image and task instruction as input and predicts future G1 kinematics. The predicted kinematics are then executed by the whole-body tracker in MuJoCo, where we evaluate closed-loop task success. We use a held-out validation set comprising 10\% of the generated data. We evaluate 1000 rollouts for each task mode, using the initial pose from each validation sequence as the test initial condition. Success is determined by task-specific heuristic rules for each mode, with details provided in Appendix~\ref{app:metric_for_sim_eval}. Table~\ref{tab:eval_system} reports the resulting success rates.

As shown in Table~\ref{tab:eval_system}, the system achieves high success rates on object-directed navigation and floor-level box manipulation in both simulation and the real world. Surface-level interactions are more challenging, partly because the retargeted interaction data provides limited coverage of box manipulation at different support-surface heights, leading to less reliable grasps and placements.

\begin{table*}[t]
\centering
\caption{Full-system evaluation in MuJoCo simulation (with IsaacSim rendering) and real-world deployment. We report the number of successful trials over the total number of trials for each mode.}
\vspace{-2pt}
\resizebox{\textwidth}{!}{
\begin{tabular}{llcccccc}
\toprule
\textbf{Setting} & \textbf{Scene} & \textbf{Walk To} & \textbf{Turn Around} & \textbf{Pick (Floor)} & \textbf{Put (Floor)} & \textbf{Pick (Surface)} & \textbf{Put (Surface)} \\
\midrule
\multirow{2}{*}{Real-World}
  & Lab Scene        & 20/20 & 20/20 & 16/20 & 20/20 & 11/20  & 8/20  \\
  & Apartment Scene  & 19/20 & 18/20 & 18/20 & 20/20 & 13/20 & 15/20 \\
\midrule
\multirow{2}{*}{Simulation}
  & Lab Scene        & 994/1000 & 843/1000 & 731/1000 & 991/1000 & 458/1000 & 569/1000 \\
  & Apartment Scene  & 948/1000 & 813/1000 & 749/1000 & 987/1000 & 521/1000 & 722/1000 \\
\bottomrule
\end{tabular}
}
\label{tab:eval_system}
\vspace{-8mm}
\end{table*}

\vspace{-12pt}
\subsection{Effect of Synthesized Data Volume}
\label{subsec:data_scale_ablation_study}
\vspace{-8pt}

\begin{wrapfigure}[13]{r}{0.48\linewidth}   
    \vspace{-29pt}                              
    \begin{minipage}{\linewidth}
      \centering
      \includegraphics[width=\linewidth]{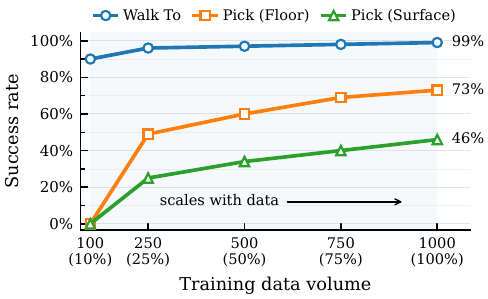} 
      \caption{Ablation of training-data volume in the lab scene. We vary the number of synthesized trajectories per mode per layout and report success rates on the lab-scene validation set.}
      \label{fig:ablation-data-volume}
    \end{minipage}
    \vspace{-8pt}
\end{wrapfigure}
\paragraph{How does the amount of synthesized training data affect performance across different tasks?}
For each layout in the lab scene, our default setting generates 1000 trajectories per mode. To study 
the effect of synthesized training-data volume, we train VLK policies using different fractions of the full dataset
and 
evaluate each policy in closed-loop MuJoCo simulation (with IsaacSim rendering) on the lab-scene validation set.
The per-mode success rates are reported in Figure~\ref{fig:ablation-data-volume}.
As shown in the figure, increasing the amount of synthesized training data consistently improves performance across the evaluated modes. Object-directed navigation achieves high success rates even with limited data, suggesting that navigation is easier to learn from synthetic supervision. In contrast, manipulation tasks benefit substantially from additional trajectories, with Pick (Surface) improving from 0\% success at 10\% data to 46\% success using the full dataset. These results indicate that contact-rich, perception-conditioned manipulation requires more scene-grounded supervision than navigation, highlighting the value of increasing synthesized data volume.

\begin{figure*}[t]
    \vspace{-20pt}
    \centering
    \includegraphics[width=\textwidth]{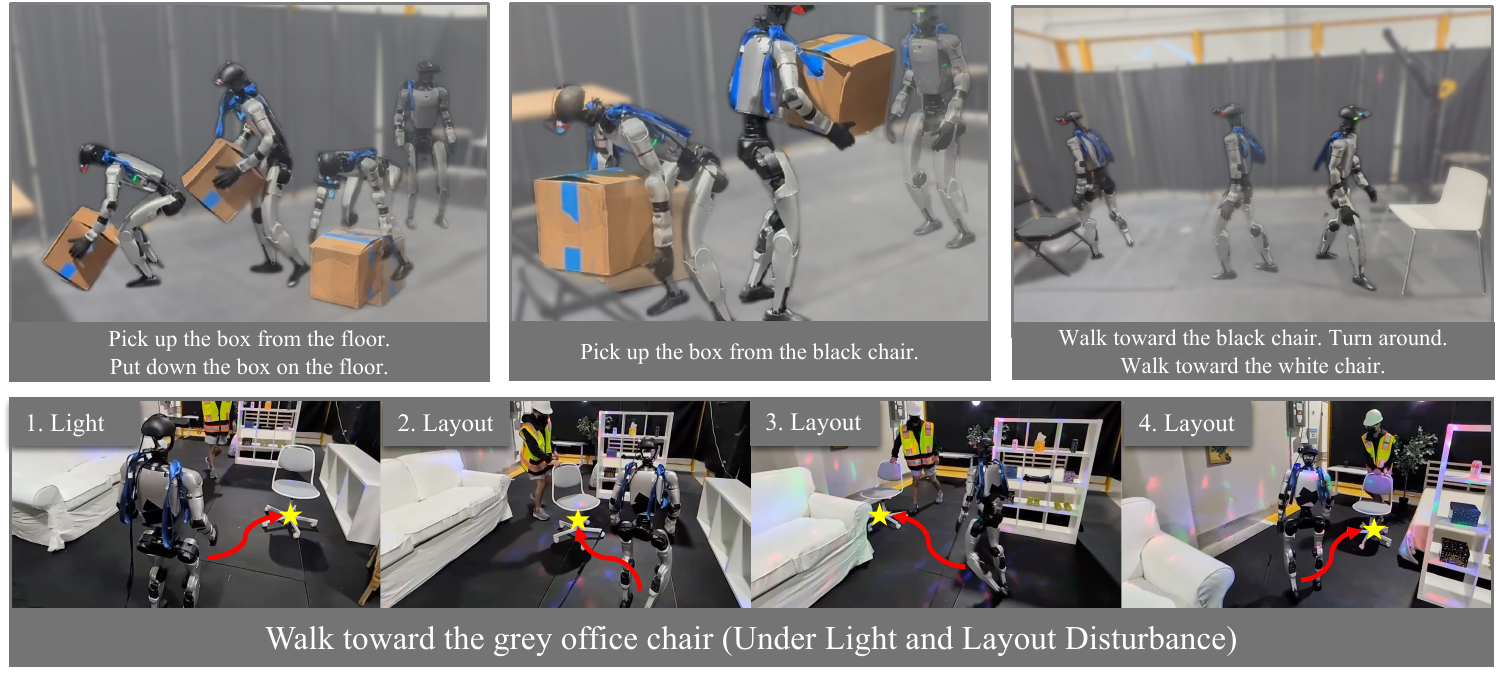}
   \vspace{-4mm}
\caption{
\textbf{Real-world deployment in lab and apartment scenes.}
The VLK system executes language-conditioned navigation and box manipulation under different layouts and visual variations.
Top: sequential lab-scene executions for manipulation and navigation.
Bottom: apartment-scene deployments under lighting (disco light) and layout variations.
}
    \label{fig:real_world_res}
    \vspace{-6mm}
\end{figure*}

 \begin{wrapfigure}[10]{r}{0.45\linewidth}
    \vspace{-14pt}
    \captionof{table}{\small Effect of visual domain randomization on walking-mode success. Evaluation is conducted in the lab scene under visual perturbations, including lighting and camera variation.}
    \vspace{-2pt}
    \small
    \begin{tabular}{lc}
      \toprule
      \textbf{Configuration} & \textbf{Success Rate (\%)} $\uparrow$ \\
      \midrule
      No randomization          & 41 \\
      Camera randomization & 48 \\
      Lighting randomization  & 87 \\
      Full randomization & 90 \\
      \bottomrule
    \end{tabular}
    \label{tab:dr_ablation}
    \vspace{-4pt}
\end{wrapfigure}

\vspace{-8pt}


\subsection{Effect of Visual Domain Randomization}
\label{subsec:domain_rand_rendering}
\vspace{-8pt}

\paragraph{How does visual domain randomization help bridge the visual sim-to-real gap?}
We evaluate whether visual domain randomization during rendering improves robustness to visual perturbations in MuJoCo simulation (with IsaacSim rendering). Specifically, we evaluate walking-mode success in the lab scene under randomized lighting and camera conditions, and compare four training configurations: no randomization, lighting randomization only, camera randomization only, and full randomization. The full setting combines lighting variation and camera variation.

As shown in Table~\ref{tab:dr_ablation}, removing visual domain randomization substantially reduces walking success from 90\% to 41\%, indicating that visual variation is important for transferring policies trained on rendered egocentric observations to the real robot. The lighting-only and camera-only settings provide a coarse ablation of appearance variation and viewpoint/calibration robustness. 
Additional real-world deployment examples are shown in Figure~\ref{fig:real_world_res}, illustrating the system operating under different lighting conditions and scene layouts.

\vspace{-10pt}

\section{Conclusion}
\vspace{-10pt}
We presented a perception-based humanoid loco-manipulation system trained from synthesized interactions in reconstructed 3D scenes. Our pipeline generates Unitree G1-compatible navigation and box-manipulation trajectories, renders paired egocentric observations, and trains a VLK policy to predict whole-body kinematic trajectories for tracker-based execution. Experiments in simulation and on the physical Unitree G1 demonstrate object-directed navigation and box manipulation, showing that scene-grounded synthetic supervision can support sim-to-real humanoid loco-manipulation. These results suggest that synthesizing robot-compatible interactions in reconstructed scenes is a promising path toward scalable perception-conditioned humanoid control.

\vspace{-3mm}
\paragraph{Limitations.}
Our current interaction synthesis is limited by the coverage of OMOMO, which contains interactions with a limited set of large objects. As a result, the generated behaviors are better suited to bimanual transport of box-like objects than to grasping small objects such as cups or tools. Similarly, the contact-aware tracker stabilizes large-object manipulation through wrist-object contact, but does not address precise object grasping. Extending VLK to small-object manipulation would require richer interaction data and a low-level controller designed for precise object grasping.




\acknowledgments{We thank Xiaoyu Huang for preparing a clean retargeted OMOMO dataset, Alejandro Escontrela for guidance on 3DGS training, Jenny Zhang and Hunter Liu for providing realistic box assets for rendering, and Eric Lalumiere for providing the camera mount. We also thank Zhen Wu, Shibo Zhao, and Yunsheng Tian for valuable discussions on motion synthesis and Real2Sim pipelines. We are grateful to Haozhi Qi, Linda Shih, Youjian Huang, and Roberto Ceja for their support with the experimental hardware, and to Shuying Deng, Zihan Wang, and Charlie Cheng for their valuable feedback on the design and visual presentation of the paper figures and video.}


\bibliography{main}  

@article{ma2025running,
  title={Running vlas at real-time speed},
  author={Ma, Yunchao and Zhou, Yizhuang and Yang, Yunhuan and Wang, Tiancai and Fan, Haoqiang},
  journal={arXiv preprint arXiv:2510.26742},
  year={2025}
}

@misc{liao2025beyondmimic,
      title={BeyondMimic: From Motion Tracking to Versatile Humanoid Control via Guided Diffusion}, 
      author={Qiayuan Liao and Takara E. Truong and Xiaoyu Huang and Yuman Gao and Guy Tevet and Koushil Sreenath and C. Karen Liu},
      year={2025},
      eprint={2508.08241},
      archivePrefix={arXiv},
      primaryClass={cs.RO},
      url={https://arxiv.org/abs/2508.08241}
}

@inproceedings{brohan2023rt2,
  title     = {{RT-2}: Vision-Language-Action Models Transfer Web Knowledge to Robotic Control},
  author    = {Zitkovich, Brianna and Yu, Tianhe and Xu, Sichun and Xu, Peng and Xiao, Ted and Xia, Fei and Wu, Jialin and Wohlhart, Paul and Welker, Stefan and Wahid, Ayzaan and Vuong, Quan and Vanhoucke, Vincent and Tran, Huong and Soricut, Radu and Singh, Anikait and Singh, Jaspiar and Sermanet, Pierre and Sanketi, Pannag R. and Salazar, Grecia and Ryoo, Michael S. and Reymann, Krista and Rao, Kanishka and Pertsch, Karl and Mordatch, Igor and Michalewski, Henryk and Lu, Yao and Levine, Sergey and Lee, Lisa and Lee, Tsang-Wei Edward and Leal, Isabel and Kuang, Yuheng and Kalashnikov, Dmitry and Julian, Ryan and Joshi, Nikhil J. and Irpan, Alex and Ichter, Brian and Hsu, Jasmine and Herzog, Alexander and Hausman, Karol and Gopalakrishnan, Keerthana and Fu, Chuyuan and Florence, Pete and Finn, Chelsea and Dubey, Kumar Avinava and Driess, Danny and Ding, Tianli and Choromanski, Krzysztof Marcin and Chen, Xi and Chebotar, Yevgen and Carbajal, Justice and Brown, Noah and Brohan, Anthony and Arenas, Montserrat Gonzalez and Han, Kehang},
  booktitle = {Proceedings of The 7th Conference on Robot Learning},
  pages     = {2165--2183},
  year      = {2023},
  volume    = {229},
  series    = {Proceedings of Machine Learning Research},
  publisher = {PMLR}
}

@inproceedings{kim2024openvla,
  title     = {{OpenVLA}: An Open-Source Vision-Language-Action Model},
  author    = {Kim, Moo Jin and Pertsch, Karl and Karamcheti, Siddharth and Xiao, Ted and Balakrishna, Ashwin and Nair, Suraj and Rafailov, Rafael and Foster, Ethan P. and Sanketi, Pannag R. and Vuong, Quan and Kollar, Thomas and Burchfiel, Benjamin and Tedrake, Russ and Sadigh, Dorsa and Levine, Sergey and Liang, Percy and Finn, Chelsea},
  booktitle = {Proceedings of The 8th Conference on Robot Learning},
  pages     = {2679--2713},
  year      = {2025},
  volume    = {270},
  series    = {Proceedings of Machine Learning Research},
  publisher = {PMLR}
}

@inproceedings{black2024pi0,
  title     = {{$\pi_0$}: A Vision-Language-Action Flow Model for General Robot Control},
  author    = {Black, Kevin and Brown, Noah and Driess, Danny and Esmail, Adnan and Equi, Michael and Finn, Chelsea and Fusai, Niccolo and Groom, Lachy and Hausman, Karol and Ichter, Brian and Jakubczak, Szymon and Jones, Tim and Ke, Liyiming and Levine, Sergey and Li-Bell, Adrian and Mothukuri, Mohith and Nair, Suraj and Pertsch, Karl and Shi, Lucy Xiaoyang and Tanner, James and Vuong, Quan and Walling, Anna and Wang, Haohuan and Zhilinsky, Ury},
  booktitle = {Robotics: Science and Systems},
  year      = {2025}
}

@article{bjorck2025groot,
  title   = {{GR00T N1}: An Open Foundation Model for Generalist Humanoid Robots},
  author  = {{NVIDIA} and Bjorck, Johan and Casta{\~n}eda, Fernando and Cherniadev, Nikita and Da, Xingye and Ding, Runyu and Fan, Linxi and Fang, Yu and Fox, Dieter and Hu, Fengyuan and others},
  journal = {arXiv preprint arXiv:2503.14734},
  year    = {2025}
}

@article{xue2025leverb,
  title   = {{LeVERB}: Humanoid Whole-Body Control with Latent Vision-Language Instruction},
  author  = {Xue, Haoru and Huang, Xiaoyu and Niu, Dantong and Liao, Qiayuan and Kragerud, Thomas and Gravdahl, Jan Tommy and Peng, Xue Bin and Shi, Guanya and Darrell, Trevor and Sreenath, Koushil and Sastry, Shankar},
  journal = {arXiv preprint arXiv:2506.13751},
  year    = {2025}
}

@inproceedings{jiang2025wholebodyvla,
  title     = {{WholeBodyVLA}: Towards Unified Latent {VLA} for Whole-Body Loco-Manipulation Control},
  author    = {Jiang, Haoran and Chen, Jin and Bu, Qingwen and Chen, Li and Shi, Modi and Zhang, Yanjie and Li, Delong and Suo, Chuanzhe and Wang, Chuang and Peng, Zhihui and Li, Hongyang},
  booktitle = {International Conference on Learning Representations},
  year      = {2026}
}

@inproceedings{wei2026psi0,
  title     = {{$\Psi_0$}: An Open Foundation Model Towards Universal Humanoid Loco-Manipulation},
  author    = {Wei, Songlin and Jing, Hongyi and Li, Boqian and Zhao, Zhenyu and Mao, Jiageng and Ni, Zhenhao and He, Sicheng and Liu, Jie and Liu, Xiawei and Kang, Kaidi and Zang, Sheng and Yuan, Weiduo and Pavone, Marco and Huang, Di and Wang, Yue},
  booktitle = {Robotics: Science and Systems},
  year      = {2026}
}

@inproceedings{robotwin,
  title     = {{RoboTwin}: Dual-Arm Robot Benchmark with Generative Digital Twins},
  author    = {Mu, Yao and Chen, Tianxing and Chen, Zanxin and Peng, Shijia and Lan, Zhiqian and Gao, Zeyu and Liang, Zhixuan and Yu, Qiaojun and Zou, Yude and Xu, Mingkun and Lin, Lunkai and Xie, Zhiqiang and Ding, Mingyu and Luo, Ping},
  booktitle = {Proceedings of the IEEE/CVF Conference on Computer Vision and Pattern Recognition},
  year      = {2025}
}

@article{splatsim,
  title   = {{SplatSim}: Zero-Shot Sim2Real Transfer of {RGB} Manipulation Policies Using Gaussian Splatting},
  author  = {Qureshi, Mohammad Nomaan and Garg, Sparsh and Yandun, Francisco and Held, David and Kantor, George and Silwal, Abhisesh},
  journal = {arXiv preprint arXiv:2409.10161},
  year    = {2024}
}

@article{robogsim,
  title   = {{RoboGSim}: A Real2Sim2Real Robotic Gaussian Splatting Simulator},
  author  = {Li, Xinhai and Li, Jialin and Zhang, Ziheng and Zhang, Rui and Jia, Fan and Wang, Tiancai and Fan, Haoqiang and Tseng, Kuo-Kun and Wang, Ruiping},
  journal = {arXiv preprint arXiv:2411.11839},
  year    = {2024}
}

@inproceedings{rlgsbridge,
  title     = {{RL-GSBridge}: 3D Gaussian Splatting Based Real2Sim2Real Method for Robotic Manipulation Learning},
  author    = {Wu, Yuxuan and Pan, Lei and Wu, Wenhua and Wang, Guangming and Miao, Yanzi and Wang, Hesheng},
  booktitle = {IEEE International Conference on Robotics and Automation},
  year      = {2025}
}

@inproceedings{gaussgym,
  title     = {{GaussGym}: An Open-Source Real-to-Sim Framework for Learning Locomotion from Pixels},
  author    = {Escontrela, Alejandro and Kerr, Justin and Allshire, Arthur and Frey, Jonas and Duan, Rocky and Sferrazza, Carmelo and Abbeel, Pieter},
  booktitle = {International Conference on Learning Representations},
  year      = {2026}
}

@inproceedings{yang2025omniretarget,
  title     = {{OmniRetarget}: Interaction-Preserving Data Generation for Humanoid Whole-Body Loco-Manipulation and Scene Interaction},
  author    = {Yang, Lujie and Huang, Xiaoyu and Wu, Zhen and Kanazawa, Angjoo and Abbeel, Pieter and Sferrazza, Carmelo and Liu, C. Karen and Duan, Rocky and Shi, Guanya},
  booktitle = {IEEE International Conference on Robotics and Automation},
  year      = {2026}
}

@inproceedings{araujo2025gmr,
  title     = {Retargeting Matters: General Motion Retargeting for Humanoid Motion Tracking},
  author    = {Araujo, Joao Pedro and Ze, Yanjie and Xu, Pei and Wu, Jiajun and Liu, C. Karen},
  booktitle = {IEEE International Conference on Robotics and Automation},
  year      = {2026}
}

@article{truong2025beyondmimic,
  title   = {{BeyondMimic}: From Motion Tracking to Versatile Humanoid Control via Guided Diffusion},
  author  = {Liao, Qiayuan and Truong, Takara E. and Huang, Xiaoyu and Tevet, Guy and Sreenath, Koushil and Liu, C. Karen},
  journal = {arXiv preprint arXiv:2508.08241},
  year    = {2025}
}

@article{dreamcontrol,
  title   = {{DreamControl}: Human-Inspired Whole-Body Humanoid Control for Scene Interaction via Guided Diffusion},
  author  = {Kalaria, Dvij and Harithas, Sudarshan S. and Katara, Pushkal and Kwak, Sangkyung and Bhagat, Sarthak and Sastry, Shankar and Sridhar, Srinath and Vemprala, Sai and Kapoor, Ashish and Huang, Jonathan Chung-Kuan},
  journal = {arXiv preprint arXiv:2509.14353},
  year    = {2025}
}

@article{zhang2026wholebodylocomotion,
  title   = {Learning Whole-Body Humanoid Locomotion via Motion Generation and Motion Tracking},
  author  = {Zhang, Zewei and Wen, Kehan and Xu, Michael and He, Junzhe and Li, Chenhao and Miki, Takahiro and Schwarke, Clemens and Zhang, Chong and Peng, Xue Bin and Hutter, Marco},
  journal = {arXiv preprint arXiv:2604.17335},
  year    = {2026}
}

@article{li2026fromw1,
  title   = {{FRoM-W1}: Towards General Humanoid Whole-Body Control with Language Instructions},
  author  = {Li, Peng and Zhuang, Zihan and Gao, Yangfan and Dong, Yi and Li, Sixian and Jiang, Changhao and Dou, Shihan and Xi, Zhiheng and Zhou, Enyu and Huang, Jixuan and Li, Hui and Gong, Jingjing and Ma, Xingjun and Gui, Tao and Wu, Zuxuan and Zhang, Qi and Huang, Xuanjing and Jiang, Yu-Gang and Qiu, Xipeng},
  journal = {arXiv preprint arXiv:2601.12799},
  year    = {2026}
}

@inproceedings{shao2025langwbc,
  title     = {{LangWBC}: Language-Directed Humanoid Whole-Body Control via End-to-End Learning},
  author    = {Shao, Yiyang and Huang, Xiaoyu and Zhang, Bike and Liao, Qiayuan and Gao, Yuman and Chi, Yufeng and Li, Zhongyu and Shao, Sophia and Sreenath, Koushil},
  booktitle = {Robotics: Science and Systems},
  year      = {2025}
}

@inproceedings{li2024chois,
  title     = {Controllable Human-Object Interaction Synthesis},
  author    = {Li, Jiaman and Clegg, Alexander and Mottaghi, Roozbeh and Wu, Jiajun and Puig, Xavier and Liu, C. Karen},
  booktitle = {European Conference on Computer Vision},
  year      = {2024}
}

@inproceedings{wu2025hihi,
  title     = {Human-Object Interaction from Human-Level Instructions},
  author    = {Wu, Zhen and Li, Jiaman and Xu, Pei and Liu, C. Karen},
  booktitle = {Proceedings of the IEEE/CVF International Conference on Computer Vision},
  year      = {2025}
}

@article{zhao2025resmimic,
  title   = {ResMimic: From General Motion Tracking to Humanoid Whole-Body Loco-Manipulation via Residual Learning},
  author  = {Zhao, Siheng and Ze, Yanjie and Wang, Yue and Liu, C. Karen and Abbeel, Pieter and Shi, Guanya and Duan, Rocky},
  journal = {arXiv preprint arXiv:2510.05070},
  year    = {2025}
}

@article{chen2025head,
  title   = {Hand-Eye Autonomous Delivery: Learning Humanoid Navigation, Locomotion and Reaching},
  author  = {Chen, Sirui and Ye, Yufei and Cao, Zi-Ang and Lew, Jennifer and Xu, Pei and Liu, C. Karen},
  journal = {arXiv preprint arXiv:2508.03068},
  year    = {2025}
}

@inproceedings{he2024omnih2o,
  title     = {{OmniH2O}: Universal and Dexterous Human-to-Humanoid Whole-Body Teleoperation and Learning},
  author    = {He, Tairan and Luo, Zhengyi and He, Xialin and Xiao, Wenli and Zhang, Chong and Zhang, Weinan and Kitani, Kris and Liu, Changliu and Shi, Guanya},
  booktitle = {Proceedings of The 8th Conference on Robot Learning},
  year      = {2024},
  series    = {Proceedings of Machine Learning Research},
  publisher = {PMLR}
}

@article{chen2025gmt,
  title   = {{GMT}: General Motion Tracking for Humanoid Whole-Body Control},
  author  = {Chen, Zixuan and Ji, Mazeyu and Cheng, Xuxin and Peng, Xuanbin and Peng, Xue Bin and Wang, Xiaolong},
  journal = {arXiv preprint arXiv:2506.14770},
  year    = {2025}
}

@misc{bones2025seed,
  title        = {{BONES-SEED}: Skeletal Everyday Embodiment Dataset},
  author       = {{Bones Studio}},
  year         = {2025},
  howpublished = {\url{https://huggingface.co/datasets/bones-studio/seed}},
  note         = {Hugging Face dataset. Accessed: 2026-05-22}
}

@article{li2023omomo,
  title   = {Object Motion Guided Human Motion Synthesis},
  author  = {Li, Jiaman and Wu, Jiajun and Liu, C. Karen},
  journal = {ACM Transactions on Graphics},
  volume  = {42},
  number  = {6},
  pages   = {225:1--225:11},
  year    = {2023},
  doi     = {10.1145/3618333}
}

@inproceedings{prokudin2019efficient,
  title     = {Efficient Learning on Point Clouds with Basis Point Sets},
  author    = {Prokudin, Sergey and Lassner, Christoph and Romero, Javier},
  booktitle = {Proceedings of the IEEE/CVF International Conference on Computer Vision},
  pages     = {4332--4341},
  year      = {2019}
}

@article{3dgs,
  title={3d gaussian splatting for real-time radiance field rendering.},
  author={Kerbl, Bernhard and Kopanas, Georgios and Leimk{\"u}hler, Thomas and Drettakis, George and others},
  journal={ACM Trans. Graph.},
  volume={42},
  number={4},
  pages={139--1},
  year={2023}
}

@inproceedings{hihi,
  title={Human-object interaction from human-level instructions},
  author={Wu, Zhen and Li, Jiaman and Xu, Pei and Liu, C Karen},
  booktitle={Proceedings of the IEEE/CVF International Conference on Computer Vision},
  pages={11176--11186},
  year={2025}
}

@article{intelligence2025pi_,
  title={{$\pi_0.5$}: a Vision-Language-Action Model with Open-World Generalization},
  author={Intelligence, Physical and Black, Kevin and Brown, Noah and Darpinian, James and Dhabalia, Karan and Driess, Danny and Esmail, Adnan and Equi, Michael and Finn, Chelsea and Fusai, Niccolo and others},
  journal={arXiv preprint arXiv:2504.16054},
  year={2025}
}

@misc{chip,
      title={CHIP: Adaptive Compliance for Humanoid Control through Hindsight Perturbation}, 
      author={Sirui Chen and Zi-ang Cao and Zhengyi Luo and Fernando Castañeda and Chenran Li and Tingwu Wang and Ye Yuan and Linxi "Jim" Fan and C. Karen Liu and Yuke Zhu},
      year={2026},
      eprint={2512.14689},
      archivePrefix={arXiv},
      primaryClass={cs.RO},
      url={https://arxiv.org/abs/2512.14689}, 
}

@article{ze2025twist2,
  title   = {TWIST2: Scalable, Portable, and Holistic Humanoid Data Collection System},
  author  = {Ze, Yanjie and Zhao, Siheng and Wang, Weizhuo and Kanazawa, Angjoo and Duan, Rocky and Abbeel, Pieter and Shi, Guanya and Wu, Jiajun and Liu, C. Karen},
  journal = {arXiv preprint arXiv:2511.02832},
  year    = {2025},
  url     = {https://arxiv.org/abs/2511.02832}
}

@article{li2025clone,
  title   = {CLONE: Closed-Loop Whole-Body Humanoid Teleoperation for Long-Horizon Tasks},
  author  = {Li, Yixuan and Lin, Yutang and Cui, Jieming and Liu, Tengyu and Liang, Wei and Zhu, Yixin and Huang, Siyuan},
  journal = {arXiv preprint arXiv:2506.08931},
  year    = {2025},
  url     = {https://arxiv.org/abs/2506.08931}
}

@inproceedings{zhou2019continuity,
  title     = {On the Continuity of Rotation Representations in Neural Networks},
  author    = {Zhou, Yi and Barnes, Connelly and Lu, Jingwan and Yang, Jimei and Li, Hao},
  booktitle = {Proceedings of the IEEE/CVF Conference on Computer Vision and Pattern Recognition (CVPR)},
  year      = {2019},
  pages     = {5745--5753}
}

@article{ding2025humanoid,
  title={Humanoid-vla: Towards universal humanoid control with visual integration},
  author={Ding, Pengxiang and Ma, Jianfei and Tong, Xinyang and Zou, Binghong and Luo, Xinxin and Fan, Yiguo and Wang, Ting and Lu, Hongchao and Mo, Panzhong and Liu, Jinxin and others},
  journal={arXiv preprint arXiv:2502.14795},
  year={2025}
}

@article{kalaria2025dreamcontrol,
  title={Dreamcontrol: Human-inspired whole-body humanoid control for scene interaction via guided diffusion},
  author={Kalaria, Dvij and Harithas, Sudarshan S and Katara, Pushkal and Kwak, Sangkyung and Bhagat, Sarthak and Sastry, Shankar and Sridhar, Srinath and Vemprala, Sai and Kapoor, Ashish and Huang, Jonathan Chung-Kuan},
  journal={arXiv preprint arXiv:2509.14353},
  year={2025}
}

@misc{chen2026scenebot,
  title         = {SceneBot: Contact-Prompted General Humanoid Whole Body Tracking with Scene-Interaction},
  author        = {Sirui Chen and Shibo Zhao and Zhen Wu and Jiaman Li and Guanya Shi and C. Karen Liu},
  year          = {2026},
  eprint        = {2606.27581},
  archivePrefix = {arXiv},
  primaryClass  = {cs.RO},
  url           = {https://arxiv.org/abs/2606.27581},
}

@article{luo2025sonic,
  title   = {SONIC: Supersizing Motion Tracking for Natural Humanoid Whole-Body Control},
  author  = {Luo, Zhengyi and Yuan, Ye and Wang, Tingwu and Li, Chenran and Chen, Sirui and Casta{\~n}eda, Fernando and Cao, Zi-Ang and Li, Jiefeng and Minor, David and Ben, Qingwei and Da, Xingye and Ding, Runyu and Hogg, Cyrus and Song, Lina and Lim, Edy and Jeong, Eugene and He, Tairan and Xue, Haoru and Xiao, Wenli and Wang, Zi and Yuen, Simon and Kautz, Jan and Chang, Yan and Iqbal, Umar and Fan, Linxi and Zhu, Yuke},
  journal = {arXiv preprint arXiv:2511.07820},
  year    = {2025}
}

@article{rempe2026kimodo,
  title   = {Kimodo: Scaling Controllable Human Motion Generation},
  author  = {Rempe, Davis and Zanfir, Mihai and Hu, Yuan-Ting and Li, Ruilong and Li, Chenran and Tseng, Jonathan and Liu, Lingjie and Darrell, Trevor and Kanazawa, Angjoo and Kautz, Jan and Liu, C. Karen and Fidler, Sanja and Guibas, Leonidas and Saito, Shunsuke and Litany, Or and Shafir, Gil and Lin, Tsung-Yi and Tulsiani, Shubham and Efros, Alexei A. and others},
  journal = {arXiv preprint arXiv:2603.15546},
  year    = {2026},
  url     = {https://arxiv.org/abs/2603.15546}
}

@inproceedings{yi2025egoallo,
  title     = {Estimating Body and Hand Motion in an Ego-sensed World},
  author    = {Yi, Brent and Ye, Vickie and Zheng, Maya and Li, Yunqi and M{\"u}ller, Lea and Pavlakos, Georgios and Ma, Yi and Malik, Jitendra and Kanazawa, Angjoo},
  booktitle = {Proceedings of the IEEE/CVF Conference on Computer Vision and Pattern Recognition (CVPR)},
  year      = {2025}
}

@article{mittal2025isaaclab,
  title   = {Isaac Lab: A GPU-Accelerated Simulation Framework for Multi-Modal Robot Learning},
  author  = {Mittal, Mayank and Roth, Pascal and Tigue, James and Richard, Antoine and Zhang, Octi and Du, Peter and Serrano-Mu{\~n}oz, Antonio and Yao, Xinjie and Z{\"u}rbr{\"u}gg, Ren{\'e} and Rudin, Nikita and Wawrzyniak, Lukasz and Rakhsha, Milad and Denzler, Alain and Heiden, Eric and Borovicka, Ales and Ahmed, Ossama and Akinola, Iretiayo and Anwar, Abrar and Carlson, Mark T. and Feng, Ji Yuan and Garg, Animesh and Gasoto, Renato and Gulich, Lionel and Guo, Yijie and Gussert, M. and Hansen, Alex and Kulkarni, Mihir and Li, Chenran and Liu, Wei and Makoviychuk, Viktor and Malczyk, Grzegorz and Mazhar, Hammad and Moghani, Masoud and Murali, Adithyavairavan and Noseworthy, Michael and Poddubny, Alexander and Ratliff, Nathan and Rehberg, Welf and Schwarke, Clemens and Singh, Ritvik and Smith, James Latham and Tang, Bingjie and Thaker, Ruchik and Trepte, Matthew and Van Wyk, Karl and Yu, Fangzhou and Millane, Alex and Ramasamy, Vikram and Steiner, Remo and Subramanian, Sangeeta and Volk, Clemens and Chen, CY and Jawale, Neel and Kuruttukulam, Ashwin Varghese and Lin, Michael A. and Mandlekar, Ajay and Patzwaldt, Karsten and Welsh, John and Zhao, Huihua and Anes, Fatima and Lafleche, Jean-Francois and Mo{\"e}nne-Loccoz, Nicolas and Park, Soowan and Stepinski, Rob and Van Gelder, Dirk and Amevor, Chris and Carius, Jan and Chang, Jumyung and Chen, Anka He and de Heras Ciechomski, Pablo and Daviet, Gilles and Mohajerani, Mohammad and von Muralt, Julia and Reutskyy, Viktor and Sauter, Michael and Schirm, Simon and Shi, Eric L. and Terdiman, Pierre and Vilella, Kenny and Widmer, Tobias and Yeoman, Gordon and Chen, Tiffany and Grizan, Sergey and Li, Cathy and Li, Lotus and Smith, Connor and Wiltz, Rafael and Alexis, Kostas and Chang, Yan and Chu, David and Fan, Linxi and Farshidian, Farbod and Handa, Ankur and Huang, Spencer and Hutter, Marco and Narang, Yashraj and Pouya, Soha and Sheng, Shiwei and Zhu, Yuke and Macklin, Miles and Moravanszky, Adam and Reist, Philipp and Guo, Yunrong and Hoeller, David and State, Gavriel},
  journal = {arXiv preprint arXiv:2511.04831},
  year    = {2025},
  url     = {https://arxiv.org/abs/2511.04831}
}

\appendix
\newpage

\etocdepthtag.toc{appendix}
\etocsettagdepth{main}{none}
\etocsettagdepth{appendix}{subsection}

\etocsettocstyle{\section*{Appendix Contents}}{}
\tableofcontents

\section{Synthetic Data Generation Details}
\label{app:data_generation}

This appendix provides additional details on the synthetic data generation pipeline and experimental setup. Additional simulation and real-world results are shown in the supplementary video.

\subsection{Scene Annotation}
\label{app:scene_annotation}

We annotate each reconstructed scene to support scene-aware motion synthesis. Since the 3DGS representation does not provide an explicit mesh or semantic labels, we first extract a point cloud from the reconstructed scene and use it as the geometric reference for annotation. We implement an interactive annotation tool using \texttt{viser}, which allows users to load the reconstructed 3D scene, inspect it in 3D, and add task-relevant annotations.

For semantic object annotation, we manually place oriented 3D bounding boxes around objects of interest and assign each box a semantic label, such as \textit{chair}, \textit{table}, or \textit{box}. The annotation interface supports translating, scaling, and rotating each box to align it with the corresponding object in the scene. These semantic 3D boxes are used to sample target objects for navigation tasks and define task-relevant object locations for interaction synthesis.

We also annotate walkable regions in each scene. Users specify a set of points on the floor plane, and the tool connects these points to form polygonal regions. These regions define feasible areas for sampling humanoid initial poses and navigation waypoints. Together, the semantic object boxes and walkable regions provide the geometric and semantic structure needed for scene-aware motion generation.

\subsection{Waypoint, Language Generation in 3D Scenes}
\label{app:task_generation}

\paragraph{Waypoint Generation.}
We generate sparse waypoints from the annotated 3D scenes according to task-specific rules. For navigation tasks, we first sample a target object from the annotated semantic 3D bounding boxes. We then sample the initial humanoid pose and intermediate waypoints from the annotated walkable regions. The sampled poses are constrained to be collision-free and compatible with the task objective.

For \textit{walk-toward-object} tasks, we additionally require the target object to be visible from the initial egocentric viewpoint. This constraint ensures that the rendered observation contains visual evidence of the task target, making the resulting training sample visually grounded. In practice, after sampling a candidate initial pose, we check whether the target object lies within the camera field of view and is not occluded by major scene structures. If the visibility constraint is not satisfied, we resample the initial pose. The distance between consecutive waypoints is selected heuristically to ensure that the generated walking motion remains natural. For \textit{turn-around} tasks, we randomly sample an initial position that is collision-free with respect to the 3D scene and set all waypoints to the initial position, producing in-place turning motions.

For box-interaction tasks, we generate waypoints and contact timing according to the object state and the height of the supporting surface. For picking motions, the waypoints guide the humanoid to approach the box, establish wrist-object contact, and lift the box from either the floor or a support surface. For placing motions, the waypoints guide the humanoid to carry the box toward the target support surface and release it at the desired location. Since lifting from the floor, lifting from a table, and placing onto surfaces require different arm heights and contact phases, we design the wrist contact frames and desired relative wrist poses based on the surface height. This task-specific design helps the synthesized motion maintain realistic contact timing, avoid abrupt vertical motion, and align the robot wrists with the box during lifting and placement.

The final waypoint sequence provides sparse spatial guidance for the motion synthesis model. These waypoints encourage the generated humanoid motion to follow feasible paths in the reconstructed scene.

\paragraph{Language Generation.}
Language annotations are generated automatically from task metadata. Since the task type, target object, and waypoint structure are known during data generation, we use template-based rules to assign language instructions to each trajectory.

For navigation trajectories, the instruction is instantiated from the semantic label of the sampled target object. For example, if the target object is annotated as a chair, the corresponding instruction is \textit{``walk toward the chair''}. This produces language labels that are directly grounded in the scene annotations.

For object-interaction trajectories, we use task-level templates defined for each motion type. For example, box-lifting trajectories are annotated with instructions such as \textit{``lift the box from the floor''}. When the interaction target or placement location varies, the template can be instantiated with the corresponding object or support surface label, such as \textit{``place the box on the table''}. These template-based annotations provide consistent supervision for the VLK policy while avoiding the need for manual language labeling.

\subsection{Interaction Synthesis Module}
\label{app:interaction_synthesis}

We synthesize humanoid-object interaction trajectories directly in the Unitree G1 kinematic representation. Our interaction synthesis module follows the conditional DDPM formulation of CHOIS~\cite{li2024chois}, which generates coupled human-object motion conditioned on language instructions, initial states, sparse waypoints, and object geometry. We retain the Transformer-based denoising architecture, but adapt the motion representation from SMPL-based human motion to the Unitree G1 humanoid. We also introduce a G1-specific forward-kinematics loss to improve end-effector position accuracy.

For a sequence of length $T$ with $J$ G1 joints, the humanoid motion is represented as
$\mathbf{x}^{\mathrm{g1}} = [\mathbf{p}, \mathbf{R}, \sin(\mathbf{q}), \cos(\mathbf{q})]$,
where $\mathbf{p} \in \mathbb{R}^{T \times J \times 3}$ denotes global joint positions,
$\mathbf{R} \in \mathbb{R}^{T \times J \times 6}$ denotes global joint rotations in the 6D rotation representation, and
$\mathbf{q} \in \mathbb{R}^{T \times D}$ denotes the G1 joint angles. We encode joint angles as $(\sin(\mathbf{q}), \cos(\mathbf{q}))$ rather than raw angles to avoid discontinuities from angle wrap-around. The object trajectory is represented as
$\mathbf{x}^{\mathrm{obj}} = [\mathbf{o}, \mathbf{R}^{\mathrm{obj}}]$,
where $\mathbf{o} \in \mathbb{R}^{T \times 3}$ is the global object position and
$\mathbf{R}^{\mathrm{obj}} \in \mathbb{R}^{T \times 6}$ is the object rotation relative to the canonical pose of the input object geometry. The full interaction trajectory is
$\mathbf{x} = [\mathbf{x}^{\mathrm{g1}}, \mathbf{x}^{\mathrm{obj}}]$.

The model is trained as a conditional denoising diffusion model. During training, Gaussian noise is added to the clean interaction trajectory $\mathbf{x}$ at a sampled diffusion timestep, and the Transformer denoising network predicts the clean motion or denoising target conditioned on the task context. The conditioning inputs include the language instruction, initial humanoid and object states, sparse scene-derived waypoints, object geometry, desired relative wrist poses, and wrist-object contact labels. Language instructions are encoded using CLIP text features, and object geometry is encoded using a Basis Point Set (BPS) representation followed by an MLP projection. The sparse waypoints provide scene-level spatial guidance, while the wrist-pose and contact conditions specify the intended object interaction. The desired relative wrist poses are defined in the object coordinate frame. For box-lifting motions, for example, we specify the two wrists to approach opposite sides of the box with downward-facing palms. The wrist-object contact labels indicate the expected contact phase for each wrist. These interaction conditions help constrain the generated full-body motion so that the wrists reach task-relevant contact regions while the object follows a consistent trajectory.

To adapt the CHOIS-style formulation to Unitree G1, we retarget OMOMO~\cite{li2023omomo} motions from SMPL to G1 using OmniRetarget~\cite{yang2025omniretarget} and train the interaction synthesis model on the resulting G1-object trajectories. We additionally implement a differentiable G1 forward-kinematics layer, which allows geometric losses to be applied to global body and end-effector positions. This encourages the generated joint-angle trajectories to produce accurate whole-body geometry and wrist placements, which is especially important for box-interaction motions.

\subsection{Post-Processing for Generated Motion}
\label{app:motion_postprocessing}

Foot sliding and unrealistic hand-object contact are common artifacts in human motion synthesis and human-object interaction synthesis. To improve the quality of the generated motions, we apply lower-body post-processing following prior work~\cite{hihi}. The key idea is to use the predicted foot-contact labels to identify contact onsets and contact phases. During each contact phase, we constrain the corresponding robot foot to remain fixed at its contact position by solving inverse kinematics.

To correct hand-object contact artifacts, we adapt the optimization approach from EgoAllo~\cite{yi2025egoallo} and introduce a wrist-pose matching term that drives the position and rotation of each wrist toward the input wrist pose expressed in the object's local frame. This encourages more realistic two-hand grasping contact during object interaction.

\subsection{Domain Randomization in Rendering}
\label{app:domain_randomization}

To improve the real-world utility of the rendered data, we apply domain randomization during egocentric rendering. The randomization covers camera parameters, lighting, and image appearance. Camera randomization perturbs the camera extrinsics and focal length to account for small calibration mismatches between the virtual and physical ZED~2i camera setups. Lighting randomization changes the dome-light intensity and yaw rotation. We also apply image-space appearance augmentations, including perturbations to brightness, contrast, saturation, and hue, as well as Gaussian noise and Gaussian blur. We summarize the domain randomization ranges in Table~\ref{app:domain_rand_app}.

\begin{table}[t]
\centering
\small
\caption{Domain randomization parameters used for egocentric rendering.}
\label{app:domain_rand_app}
\begin{tabular}{ll}
\toprule
\textbf{Randomization} & \textbf{Range} \\
\midrule
Camera translation jitter & $\pm 2$ cm \\
Camera rotation jitter & $\pm 3^\circ$ \\
Camera focal perturbation & $\pm 2\%$ \\
Dome-light intensity & $[400, 1500]$ \\
Dome-light yaw rotation & $[-\pi, \pi]$ \\
Brightness & $[0.8, 1.2]$ \\
Contrast & $[0.8, 1.3]$ \\
Saturation & $[0.7, 1.4]$ \\
Hue & $[-0.05, 0.05]$ \\
Gaussian noise std. & $[0, 0.05]$ \\
Gaussian blur sigma & $[0, 1.0]$ \\
\bottomrule
\end{tabular}
\end{table}

\subsection{Dataset Examples}
\label{app:dataset_examples}

\begin{figure*}[t]
    \centering
    \includegraphics[width=\textwidth]{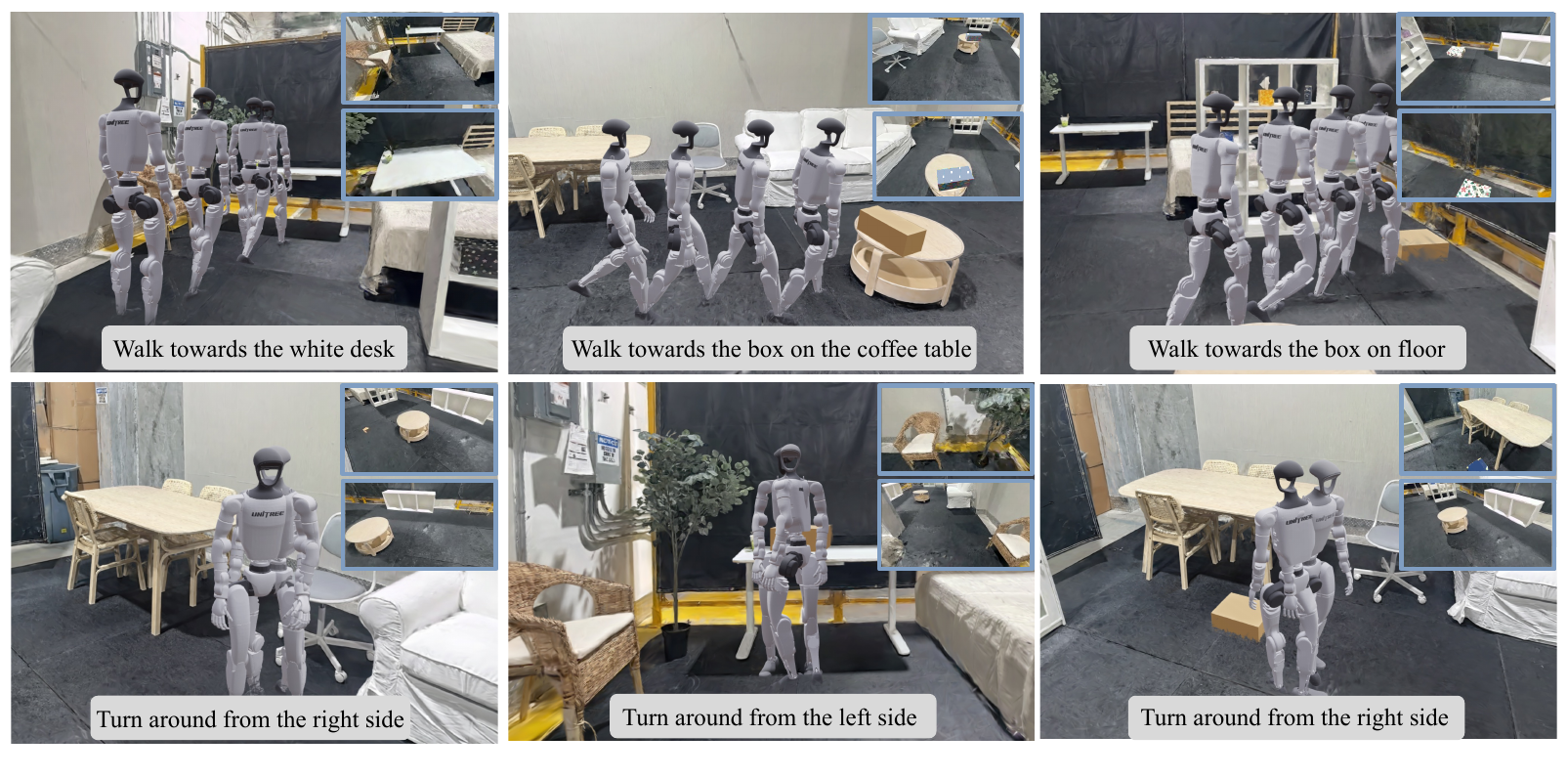}
    \caption{
    Examples from the synthetic vision-language-kinematics dataset. Each example pairs an egocentric RGB observation and language instruction with a G1 kinematic trajectory. 
    }
    \label{fig:dataset_examples}
\end{figure*}

Figure~\ref{fig:dataset_examples} shows additional representative examples from our synthetic dataset. Each example contains a rendered egocentric RGB observation, a language instruction, and the corresponding G1 kinematic trajectory generated in an annotated 3D scene. Visualizations of more sequences are provided in the supplementary video.

\section{Experiment Details}
This section presents supplementary details for model training, evaluation, and deployment. We describe the objective functions used to fine-tune the VLK policy, the metrics adopted for simulation evaluation, and the practical considerations required for real-world deployment, including motion-blur handling and system-level profiling.

\subsection{VLK Training Loss Details}
\label{app:vlk_training_details}

\paragraph{Training Objective.}
We initialize the VLK policy from a pretrained $\pi_{0.5}$~\cite{intelligence2025pi_} and fine-tune the full model on our generated vision-language-kinematics dataset. We adapt the action space of the pretrained model to our G1 kinematic representation.

We train the policy with an $x_0$-prediction objective in a flow-matching formulation. Given a ground-truth future trajectory $\mathbf{x}_{t+1:t+H}$, Gaussian noise $\boldsymbol{\epsilon}$, and interpolation coefficient $\alpha$, we construct the noisy trajectory
\[
\mathbf{x}^{\alpha}_{t+1:t+H}
=
\alpha\boldsymbol{\epsilon}
+
(1-\alpha)\mathbf{x}_{t+1:t+H}.
\]
The policy predicts the clean trajectory conditioned on the noisy trajectory, interpolation coefficient, current egocentric observation, task instruction, and current G1 kinematic state:
\[
\hat{\mathbf{x}}_{t+1:t+H}
=
\pi_\theta
(
\mathbf{x}^{\alpha}_{t+1:t+H},
\alpha,
o_t,
\ell,
\mathbf{x}_t
).
\]
The primary trajectory reconstruction loss is
\[
\mathcal{L}_{\mathrm{traj}}
=
\left\|
\hat{\mathbf{x}}_{t+1:t+H}
-
\mathbf{x}_{t+1:t+H}
\right\|_2^2.
\]

\paragraph{Auxiliary Losses.}
To improve physical consistency and execution quality, we introduce auxiliary objectives for foot-floor contact prediction, accumulated root-trajectory consistency, forward-kinematics accuracy, and foot-skating regularization.

\textbf{Foot-floor contact loss.}
We predict foot-floor contact labels to identify stance phases used by the foot-skating regularizer. Since contact events are sparse, we supervise the predicted foot-contact labels using a focal loss:
\[
\mathcal{L}_{\mathrm{foot\text{-}contact}}
=
\mathrm{FocalLoss}
(
\hat{\mathbf{m}},
\mathbf{m}
),
\]
where $\hat{\mathbf{m}}$ and $\mathbf{m}$ denote the predicted and ground-truth foot-contact labels.

\textbf{Accumulated root-trajectory loss.}
The trajectory representation parameterizes root motion using relative displacements. We integrate the predicted root displacements to recover the global root trajectory and penalize accumulated drift:
\[
\mathcal{L}_{\mathrm{acc\text{-}root}}
=
\frac{1}{H}
\sum_{k=1}^{H}
\left\|
\hat{\mathbf{p}}^{\mathrm{root}}_{t+k}
-
\mathbf{p}^{\mathrm{root}}_{t+k}
\right\|_2^2 .
\]

\textbf{Forward-kinematics loss.}
We apply forward kinematics to the predicted joint angles and supervise the resulting ankle and wrist positions:
\[
\mathcal{L}_{\mathrm{fk}}
=
\mathcal{L}_{\mathrm{fk}}^{\mathrm{ankle}}
+
\mathcal{L}_{\mathrm{fk}}^{\mathrm{wrist}},
\]
where
\[
\mathcal{L}_{\mathrm{fk}}^{\mathrm{ankle}}
=
\frac{1}{H|\mathcal{A}|}
\sum_{k=1}^{H}
\sum_{j\in\mathcal{A}}
\left\|
\hat{\mathbf{p}}^{j}_{t+k}
-
\mathbf{p}^{j}_{t+k}
\right\|_2^2,
\qquad
\mathcal{L}_{\mathrm{fk}}^{\mathrm{wrist}}
=
\frac{1}{H|\mathcal{W}|}
\sum_{k=1}^{H}
\sum_{j\in\mathcal{W}}
\left\|
\hat{\mathbf{p}}^{j}_{t+k}
-
\mathbf{p}^{j}_{t+k}
\right\|_2^2.
\]
Here, $\mathcal{A}$ and $\mathcal{W}$ denote the ankle and wrist end-effectors, respectively.

\textbf{Foot-skating regularization loss.}
To discourage unrealistic foot sliding during stance phases, we penalize horizontal foot velocity when a foot is predicted to be in contact with the ground:
\[
\mathcal{L}_{\mathrm{foot}}
=
\frac{1}{H|\mathcal{F}|}
\sum_{k=1}^{H}
\sum_{j\in\mathcal{F}}
\hat{m}^{j}_{t+k}
\left\|
\mathbf{v}^{j,xy}_{t+k}
\right\|_2^2 ,
\]
where $\mathcal{F}$ denotes the foot end-effectors, $\hat{m}^{j}_{t+k}$ is the predicted foot-contact label for foot $j$, and $\mathbf{v}^{j,xy}_{t+k}$ is the corresponding foot velocity projected onto the ground plane.

The final training objective is
\[
\mathcal{L}_{\mathrm{total}}
=
1.0\,\mathcal{L}_{\mathrm{traj}}
+
0.5\,\mathcal{L}_{\mathrm{foot\text{-}contact}}
+
0.2\,\mathcal{L}_{\mathrm{acc\text{-}root}}
+
1.0\,\mathcal{L}_{\mathrm{fk}}^{\mathrm{ankle}}
+
1.0\,\mathcal{L}_{\mathrm{fk}}^{\mathrm{wrist}}
+
0.05\,\mathcal{L}_{\mathrm{foot}}.
\]

\subsection{Metric for Evaluation in Simulation}
\label{app:metric_for_sim_eval}

While the RL tracking controller is trained in IsaacLab~\cite{mittal2025isaaclab}, we perform sim2sim evaluation in MuJoCo. During evaluation, the simulator is paused while the VLK policy performs trajectory inference at the end of each motion chunk, avoiding the need for real-time chunked execution. For each task mode, we hold out 10\% of the synthesized trajectories (1000 trajectories per mode) for evaluation. The humanoid is initialized from the initial state of each held-out trajectory and evaluated on whether it can successfully complete the corresponding language-conditioned task.

For ``Walk To'' tasks, a rollout is considered successful if the humanoid reaches the specified target object or box, stops within 0.5\,m of the target, and avoids colliding with or penetrating the target geometry.

For ``Turn Around'' tasks, success requires the humanoid to perform a clear turning motion toward the instructed direction, rather than remaining stationary or continuing forward motion.

For ``Pick Up'' tasks, the humanoid must successfully grasp a box placed either on the floor or on top of an object and maintain stable holding for at least 20 consecutive frames.

For ``Put Down'' tasks, success is defined as the humanoid reaching the specified target object and successfully placing the carried box onto the target surface or floor.



\subsection{Motion Blur Handling During Deployment}
\label{app:motion_blur_deployment_details} 

\paragraph{During training.}
We augment a subset of RGB observations with synthetic motion blur:
\[
I'
=
\mathcal{B}_{\sigma}(I),
\qquad
\sigma \sim p(\sigma),
\]
where $I$ is the original image, $\mathcal{B}_{\sigma}$ denotes the blur operator with magnitude $\sigma$, and $p(\sigma)$ is restricted to a small range to preserve semantic scene content while improving robustness to mild blur artifacts.

\paragraph{During deployment.}
We maintain an image buffer covering a $0.3\,\mathrm{s}$ window. Given buffered candidate frames
\[
\left\{
I_t^{(1)},
I_t^{(2)},
\ldots,
I_t^{(N)}
\right\},
\]
the system selects the sharpest frame according to the variance of the Laplacian:
\[
S(I)
=
\mathrm{Var}\!\left(\nabla^2 I\right),
\]
where $\nabla^2 I$ denotes the image Laplacian operator. The selected observation is
\[
I_t^\ast
=
\arg\max_{I_t^{(i)}} S\!\left(I_t^{(i)}\right).
\]
The resulting frame is then passed to the VLK policy for inference.

\subsection{System Overview and Profiling}
\label{app:deployment_profiling} 
\begin{figure}[t]
\centering
\includegraphics[width=0.95\linewidth]{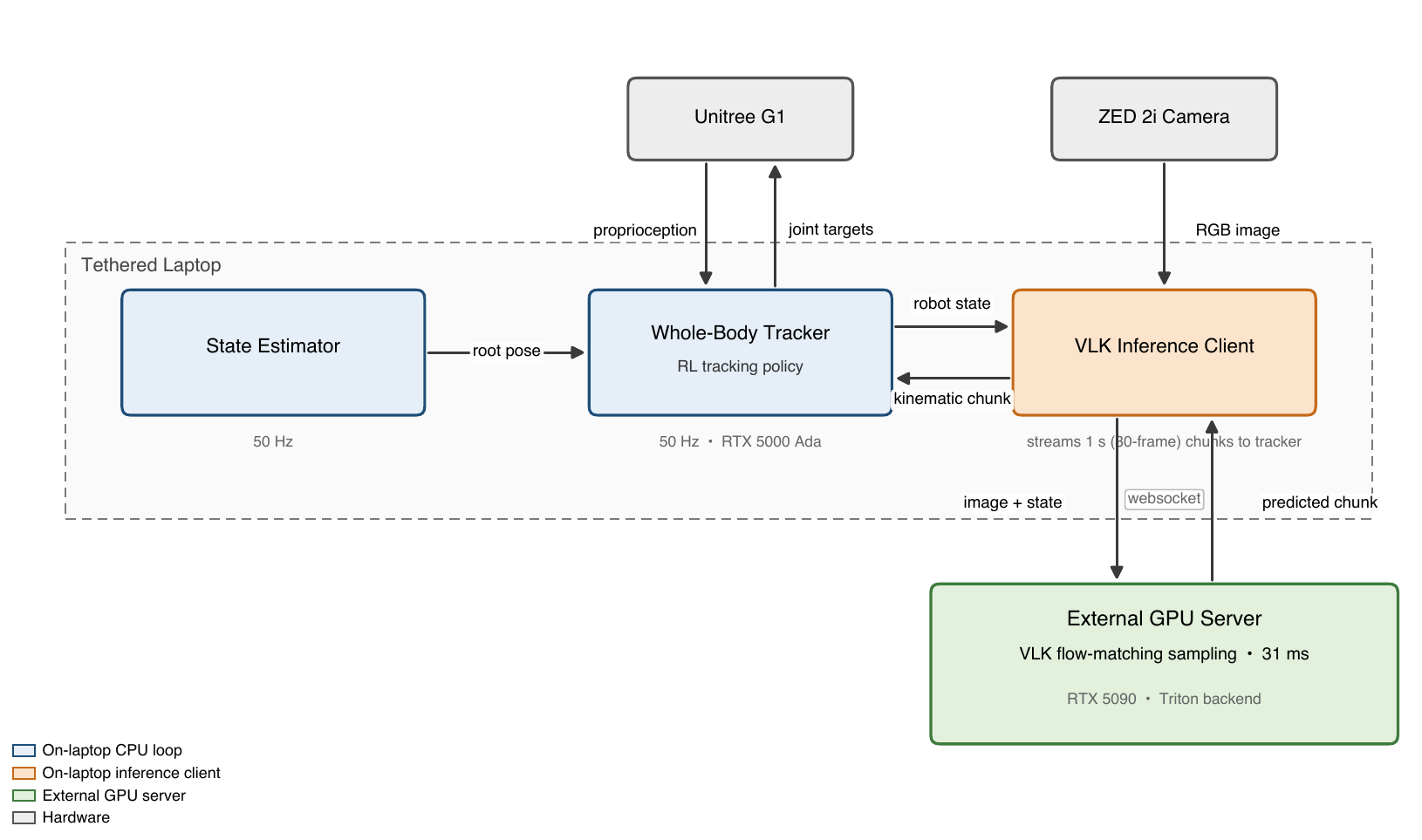}
\caption{\textbf{Deployment system overview.} Three concurrent on-robot
processes (blue: state estimation and whole-body tracking; orange: VLK
inference client) communicate through shared memory and inter-process
queues. The client offloads flow-matching sampling to a remote VLK server
(green) over a websocket. The whole-body tracker runs at $50\,\mathrm{Hz}$
and emits joint-level PD targets, while the inference loop replans at
$\sim\!1.8\,\mathrm{Hz}$; predicted kinematic chunks are spliced into the
active reference with a $10$-frame overlap.}
\label{fig:system_overview}
\end{figure}

We deploy the VLK policy on the physical Unitree G1 with a multi-process
runtime that decouples high-rate whole-body control from lower-rate
vision-language inference. Figure~\ref{fig:system_overview} shows the
architecture. Three concurrent processes run on a tethered laptop---a
state estimator, a whole-body tracker, and a VLK inference client---and
communicate through shared memory and inter-process queues. The inference
client streams the latest egocentric image and robot kinematic state to an
external GPU server, which performs the flow-matching trajectory sampling,
and merges the returned kinematic chunks into the tracker's reference stream.

\paragraph{Whole-body tracking loop.}
The tracker runs at $50\,\mathrm{Hz}$ on an RTX 5000 Ada. Each tick reads
robot proprioception, fetches the latest root pose from the state estimator,
splices any newly returned VLK kinematic chunk into the active reference,
runs the RL tracking policy, and sends joint-level PD targets to the robot.
Across our deployments, the per-tick computation averages
$4.3\,\mathrm{ms}$, with no missed deadlines observed over a full task
rollout.

\paragraph{VLK inference loop.}
A separate inference client maintains the most recent egocentric RGB frame
from the ZED 2i and the most recent robot state in shared memory. Each
replan packs the observation, sends it to the external GPU server over a
websocket, receives a predicted $H\!=\!30$-frame kinematic chunk (one second
of future motion at $30\,\mathrm{Hz}$), and converts the canonical-frame
prediction into world-frame motion references via forward kinematics.
Predicted chunks overlap by $10$ frames (Section~\ref{deployment}) and are consumed by the
$50\,\mathrm{Hz}$ tracker as a streaming reference rather than re-predicted
every control tick. On the RTX 5090, VLK inference itself takes
$31\,\mathrm{ms}$, and the end-to-end replan---including websocket transport,
input packing, and post-processing---averages $\sim\!63\,\mathrm{ms}$
(Table~\ref{tab:replan_breakdown}); since each chunk covers $1\,\mathrm{s}$
of motion, this leaves ample headroom and chunks do not accumulate.

\paragraph{Asynchronous chunk merging.}
Because chunks are produced asynchronously while the tracker runs at
$50\,\mathrm{Hz}$, a freshly arrived chunk is spliced into the active
reference with a $10$-frame overlap (Section~\ref{deployment}). The merge step contributes
$\sim\!3.4\,\mathrm{ms}$ to the tracker tick on which a new chunk lands and
early-returns otherwise, keeping the tracker comfortably within its
$50\,\mathrm{Hz}$ schedule.

\begin{table}[t]
\centering
\small
\caption{End-to-end VLK replan latency breakdown, averaged across one
deployment. The total of $\sim\!63\,\mathrm{ms}$ is well below the
$\sim\!555\,\mathrm{ms}$ replan period, leaving $\sim\!8.8\times$ headroom
against backlog formation.}
\label{tab:replan_breakdown}
\begin{tabular}{@{}lrr@{}}
\toprule
Stage & Latency (ms) & Share \\
\midrule
Image fetch from camera buffer                   &  5.4 & \phantom{0}9\% \\
Observation packing (state norm.\ $+$ image enc.) &  7.5 & 12\%           \\
Server roundtrip $+$ GPU flow-matching sampling   & 37.0 & 59\%           \\
Output denormalization                            &  4.2 & \phantom{0}7\% \\
World-frame transform $+$ forward kinematics      &  7.2 & 11\%           \\
Reference merge into tracker stream               &  3.4 & \phantom{0}5\% \\
\midrule
\textbf{Total}                                    & \textbf{63.0} & \textbf{100\%} \\
\bottomrule
\end{tabular}
\end{table}


\end{document}